\documentclass{article}

\usepackage{microtype}
\usepackage{graphicx}
\usepackage{subcaption}
\usepackage{booktabs} 

\usepackage{hyperref}

\usepackage[accepted]{my_icml2023}

\usepackage{amsmath}
\usepackage{amssymb}
\usepackage{mathtools}
\usepackage{amsthm}

\usepackage[capitalize,noabbrev]{cleveref}

\theoremstyle{plain}
\newtheorem{theorem}{Theorem}[section]
\newtheorem{proposition}[theorem]{Proposition}

\newtheorem{definition}[theorem]{Definition}

\theoremstyle{remark}

\usepackage[inline]{enumitem}
\usepackage{multirow}

\pdfoutput=1

\DeclareMathOperator*{\arginf}{arg\,inf}

\DeclareMathOperator{\lse}{min}

\newcommand{\ovk}[1]{\|#1\|_{\mathrm{ov}k}}

\DeclareMathOperator*{\argmax}{arg\,max} 
\DeclareMathOperator*{\argmin}{arg\,min} 
\DeclareMathOperator*{\argsup}{arg\,sup} 
\DeclareMathOperator{\prox}{prox} 
\DeclareMathOperator{\supp}{supp}
\DeclareMathOperator{\sign}{sign} 
\DeclareMathOperator{\arcsinh}{asinh}

\definecolor{mygreen}{rgb}{0.0, 0.7, 0.0}
\newcommand{\R}{\mathbb{R}}
\newcommand{\Pcal}{\mathcal{P}}
\newcommand{\Ccal}{\mathcal{C}}
\newcommand{\bff}{\mathbf{f}}

\newcommand{\bx}{\mathbf{x}}
\newcommand{\bg}{\mathbf{g}}

\newcommand{\by}{\mathbf{y}}
\newcommand{\bz}{\mathbf{z}}
\newcommand{\bw}{\mathbf{w}}
\newcommand{\bu}{\mathbf{u}}
\newcommand{\bv}{\mathbf{v}}

\usepackage{wrapfig, framed, caption}
\usepackage{multirow}
\hypersetup{colorlinks,linkcolor={blue},citecolor={blue},urlcolor={black}}  

\usepackage[textsize=tiny]{todonotes}

\usepackage[capitalize,noabbrev]{cleveref}

\usepackage[textsize=tiny]{todonotes}

\icmltitlerunning{Monge, Bregman and Occam}

\begin{document}

\twocolumn[
\icmltitle{Monge, Bregman and Occam:
Interpretable Optimal Transport in High-Dimensions with Feature-Sparse Maps}




\begin{icmlauthorlist}
\icmlauthor{Marco Cuturi}{app}
\icmlauthor{Michal Klein}{app}
\icmlauthor{Pierre Ablin}{app}
\end{icmlauthorlist}
\icmlaffiliation{app}{Apple}

\icmlcorrespondingauthor{Marco Cuturi}{cuturi@apple.com}

\icmlkeywords{Machine Learning, ICML}

\vskip 0.3in
]



\printAffiliationsAndNotice{}  

\begin{abstract}
Optimal transport (OT) theory focuses, among all maps $T:\R^d\rightarrow \R^d$ that can morph a probability measure onto another, on those that are the ``thriftiest'', i.e. such that the averaged cost $c(\bx, T(\bx))$ between $\bx$ and its image $T(\bx)$ be as small as possible. Many computational approaches have been proposed to estimate such \textit{Monge} maps when $c$ is the $\ell_2^2$ distance, e.g., using entropic maps~\citep{pooladian2021entropic}, or neural networks~\citep{makkuva2020optimal,korotin2020wasserstein}. We propose a new model for transport maps, built on a family of translation invariant costs $c(\bx,\by):=h(\bx-\by)$, where $h:=\tfrac{1}{2}\|\cdot\|_2^2+\tau$ and $\tau$ is a regularizer. We propose a generalization of the entropic map suitable for $h$, and highlight a surprising link tying it with the \textit{Bregman} centroids of the divergence $D_h$ generated by $h$, and the proximal operator of $\tau$. We show that choosing a sparsity-inducing norm for $\tau$ results in maps that apply \textit{Occam}'s razor to transport, in the sense that the \textit{displacement} vectors $\Delta(\bx):= T(\bx)-\bx$ they induce are sparse, with a sparsity pattern that varies depending on $\bx$. We showcase the ability of our method to estimate meaningful OT maps for high-dimensional single-cell transcription data, in the $34000$-$d$ space of gene counts for cells, \textit{without} using dimensionality reduction, thus retaining the ability to interpret all displacements at the gene level.
\end{abstract}

\section{Introduction}\label{sec:intro}
A fundamental task in machine learning is learning how to \textit{transfer} observations from a source to a target probability measure.
For such problems, optimal transport (OT)~\citep{santambrogio2015optimal} has emerged as a powerful toolbox  that can improve performance and guide theory in various settings. For instance, the computational approaches advocated in OT have been used to transfer knowledge across datasets in domain adaptation tasks~\citep{courty2016optimal,courty2017joint}, train generative models \citep{NIPS2016_728f206c,arjovsky2017wasserstein,2017-Genevay-AutoDiff,salimans2018improving}, and realign datasets in natural sciences~\citep{janati2019wasserstein,schiebinger2019optimal}.

\textbf{High-dimensional Transport.} OT finds its most straightforward and intuitive use-cases in low-dimensional geometric domains (grids and meshes, graphs, etc...). This work focuses on the more challenging problem of using it on distributions in $\R^d$, with $d\gg 1$. In $\R^d$, the ground cost $c(\bx,\by)$ between observations $\bx,\by$ is often the $\ell_2$ metric or its square $\ell_2^2$. However, when used on large-$d$ data samples, that choice is rarely meaningful. This is due to the curse-of-dimensionality associated with OT estimation~\citep{dudley1966weak,weed2017sharp} and the fact that the Euclidean distance loses its discriminative power as dimension grows. To mitigate this, practitioners rely on dimensionality reduction, either in \textit{two steps}, before running OT solvers, using, e.g., PCA, a VAE, or a sliced-Wasserstein approach~\citep{rabin2012wasserstein, bonneel2015sliced}; or \textit{jointly}, by estimating both a projection and transport, e.g., on hyperplanes~\citep{niles2022estimation,paty2019subspace,lin2020projection,pmlr-v139-huang21e,lin2021projection}, lines~\citep{deshpande2019max,kolouri2019generalized}, trees~\citep{le2019tree} or more advanced featurizers~\citep{salimans2018improving}. However, an obvious drawback of these approaches is that transport maps estimated in reduced dimensions are hard to interpret in the original space~\citep{NEURIPS2019_f9beb1e8}.

\begin{figure*}[t!]
   \centering
\begin{tabular}{c|ccc}
\multirow{ 2}{*}[3em]{
\includegraphics[width=.22\textwidth]{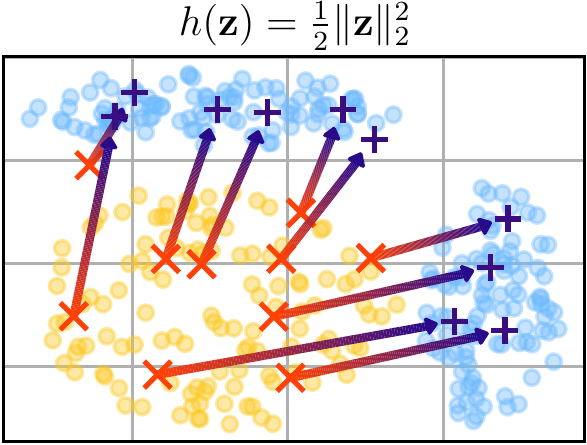}
}
&
\includegraphics[width=.22\textwidth]{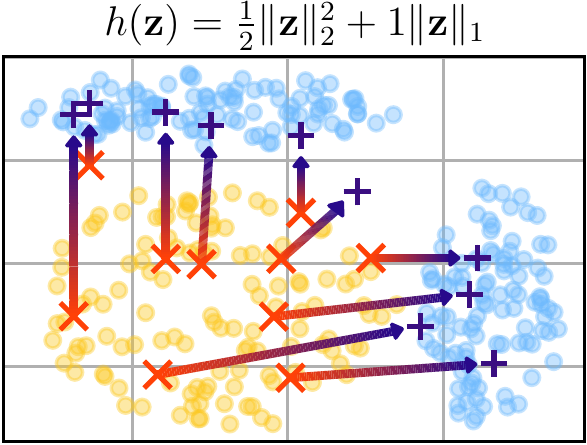}&
\includegraphics[width=.22\textwidth]{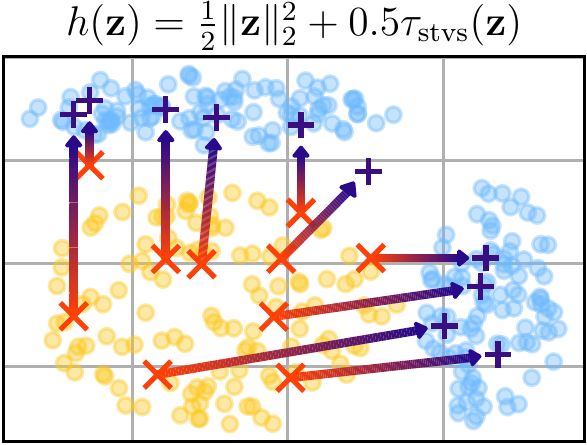}&
\includegraphics[width=.22\textwidth]{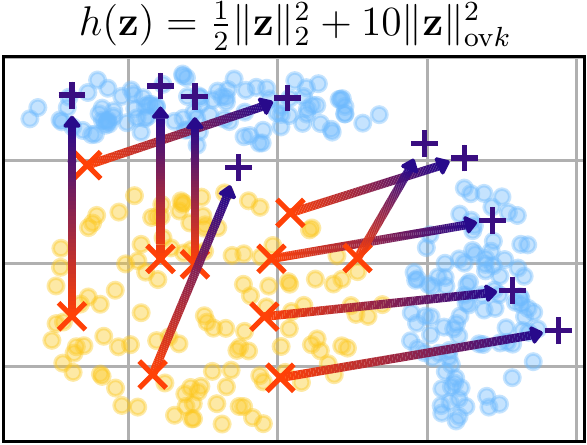}\\
&
\includegraphics[width=.22\textwidth]{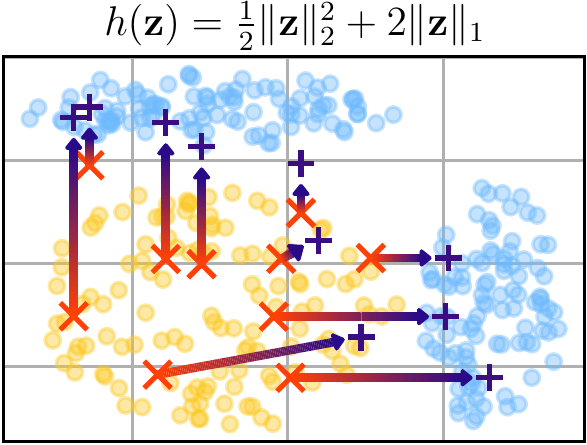}&
\includegraphics[width=.22\textwidth]{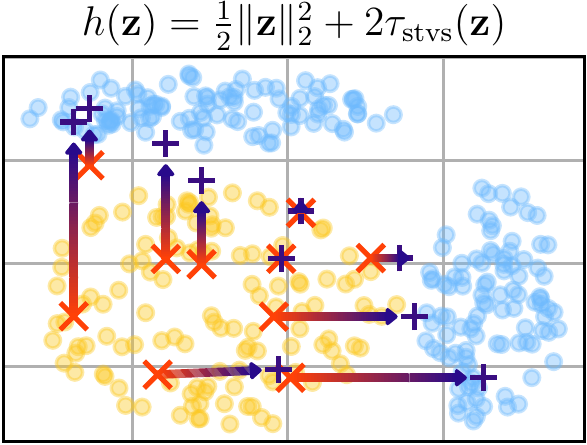}&
\includegraphics[width=.22\textwidth]{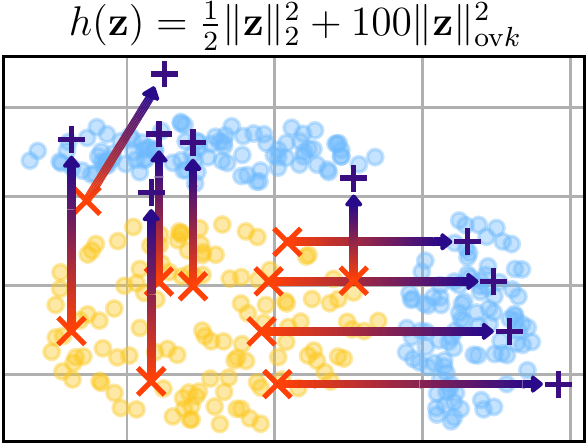}\\
\end{tabular}

\begin{subfigure}[b]{0.65\textwidth}
    \centering
    \includegraphics[width=\linewidth]{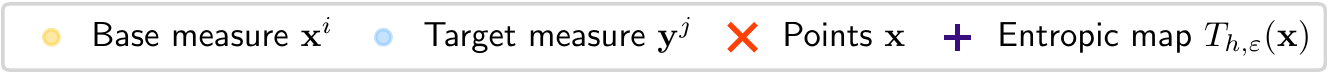}
\end{subfigure}%
    \caption{Plots of entropic map estimators $T_{h, \varepsilon}$, as defined in Prop.~\ref{prop:testimator}, to map a 2D measure supported on $(\bx^i)$ onto that supported on $(\by^j)$, for various costs $h$. The displacements $\Delta(\bx)=T_{h, \varepsilon}(\bx)-\bx$ of unseen points are displayed as \textit{arrows}. From left to right: standard $\ell_2^2$ norm,  Elastic $\ell_1$, STVS, and $k$-support costs ($k=1$). For each proposed cost, the regularization $\gamma$ is small on the top row and high on the bottom. Displacements are not sparse for the $\ell_2^2$ cost but become increasingly so as $\gamma$ grows, with a support that varies with input $\bx$. Note that Elastic $\ell_1$ and STVS tend to censor displacements as $\gamma$ grows, to the extent that they become null. In contrast, the $k$-support cost encourages sparsity but enforces displacements with at least $k$ non-zero values. See also~\autoref{fig:entropic_move_illustration} for aggregate results.}
    \label{fig:entropic_map_illustration} 
\end{figure*}

\textbf{Contributions.} To target high $d$ regimes, we introduce a radically different approach. We use the sparsity toolbox~\citep{hastie2015statistical,bach2012optimization} to build OT maps that are, \textit{adaptively} to input $\bx$, drastically simpler:
\begin{itemize}[leftmargin=.3cm,itemsep=.0cm,topsep=0cm,parsep=2pt]
\item We introduce a generalized entropic map~\citep{pooladian2021entropic} for translation invariant costs $c(\bx,\by):=h(\bx-\by)$, where $h$ is strongly convex. That entropic map $T_{h,\varepsilon}$ is defined almost everywhere (a.e.), and we show that it induces displacements $\Delta(\bx):=T(\bx)-\bx$ that can be cast as Bregman centroids, relative to the Bregman divergence generated by $h$.
\item When $h$ is an elastic-type regularizer, the sum of a strongly-convex term $\ell_2^2$ and a sparsifying norm $\tau$, we show that such centroids are obtained using the proximal operator of $\tau$. This induces \textit{sparse} displacements $\Delta(\bx)$, with a sparsity pattern that depends on $\bx$, controlled by the regularization strength set for $\tau$. To our knowledge, our formulation is the first in the computational OT literature that can produce features-wise sparse OT maps.
\item We apply our method to single-cell transcription data using two different sparsity-inducing proximal operators. We show that this approach succeeds in recovering meaningful maps in extremely high-dimension.
\end{itemize}

\textbf{Not the Usual Sparsity found in Computational OT.}
Let us emphasize that the sparsity studied in this work is unrelated, and, in fact, orthogonal, to the many references to sparsity found in the computational OT literature. Such references arise when computing an OT plan from $n$ to $m$ points, resulting in large $n\times m$ optimal coupling matrices. Such matrices are sparse when any point in the source measure is only associated to one or a few points in the target measure. Such sparsity acts at the level of \textit{samples}, and is usually a direct consequence of linear programming duality~\citep[Proposition 3.4]{PeyCut19}. It can be also encouraged with regularization~\citep{courty2016optimal,dessein2018regularized,blondel2018smooth} or constraints~\citep{BlondelLiu}. By contrast, sparsity in this work \textit{only} occurs relative to the \textit{features} of the displacement vector $\Delta(\bx)\in\R^d$, when moving a given $\bx$, i.e., $\|\Delta(\bx)\|_0\ll d$. Note, finally, that we do not use coupling matrices in this paper.

\textbf{Links to OT Theory with Degenerate Costs.}
Starting with the seminal work by~\citet{sudakov1979geometric}, who proved the \textit{existence} of Monge maps for the original 
\citeauthor{Monge1781} problem, studying non-strongly convex costs with gradient discontinuities~\citep[\S3]{santambrogio2015optimal} has been behind many key theoretical developments~\citep{ambrosio2003existence,ambrosio,evans1999differential,trudinger2001monge,carlier2010strategy,bianchini2014decomposition}. While these works have few practical implications, because they focus on the \textit{existence} of Monge maps, constructed by stitching together OT maps defined pointwise, they did, however, guide our work in the sense that they shed light on the difficulties that arise from ``flat'' norms such as $\ell_1$.  This has guided our focus in this work on elastic-type norms, which allow controlling the amount of sparsity through regularization strength, by analogy with the Lasso tradeoff where an $\ell_2^2$ loss is paired with an $\ell_1$ regularizer.


    \begin{figure*}[t!]
       \centering
    \begin{tabular}{c|ccc}
    \multirow{ 2}{*}[3em]{
    \includegraphics[width=.22\textwidth]{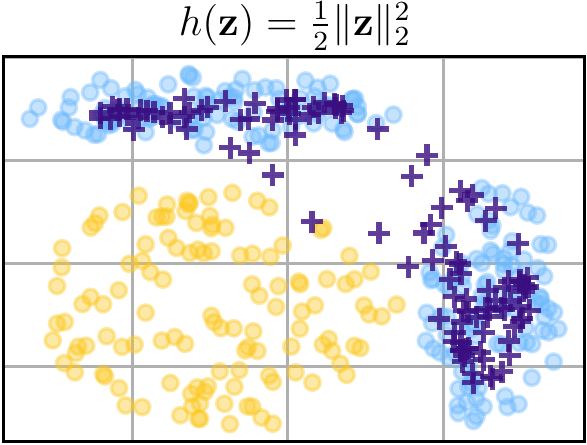}
    }
    &
    \includegraphics[width=.22\textwidth]{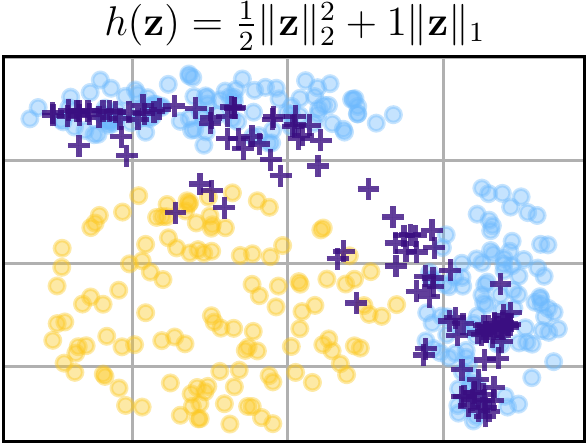}&
    \includegraphics[width=.22\textwidth]{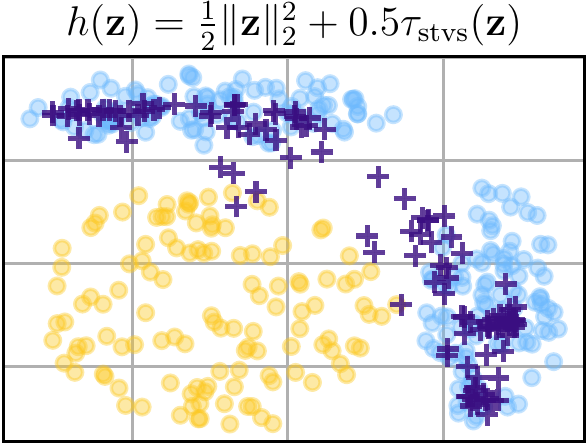}&
    \includegraphics[width=.22\textwidth]{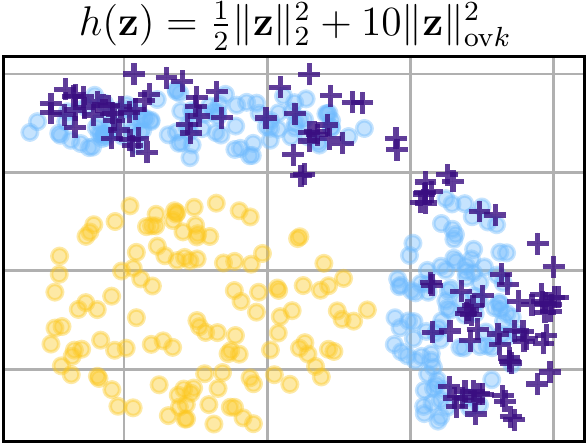}\\
    &
    \includegraphics[width=.22\textwidth]{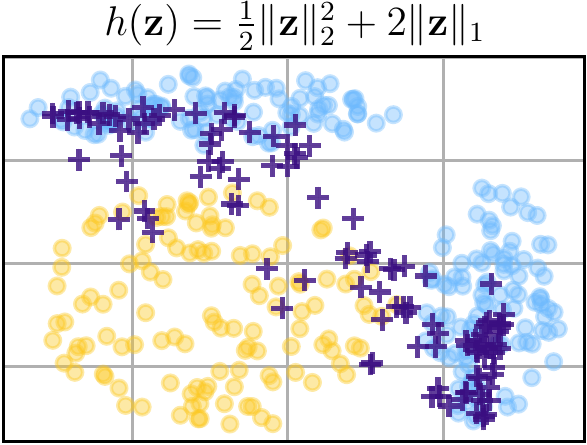}&
    \includegraphics[width=.22\textwidth]{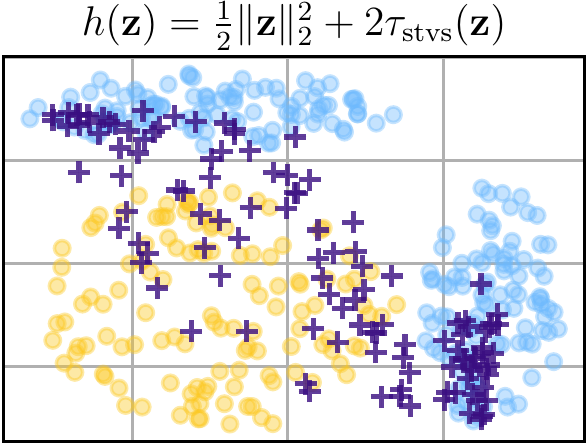}&
    \includegraphics[width=.22\textwidth]{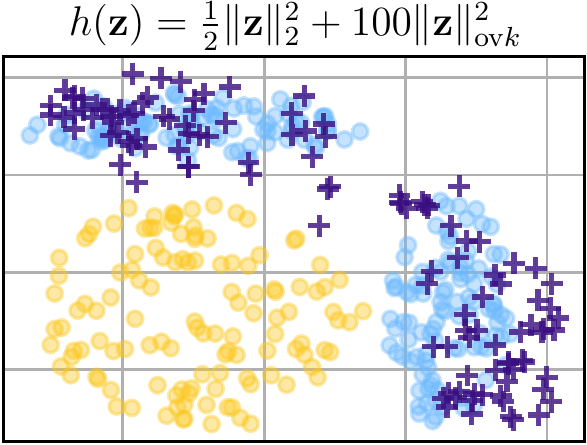}\\
    \end{tabular}
    
    \begin{subfigure}[b]{0.65\textwidth}
        \centering
        \includegraphics[width=\linewidth]{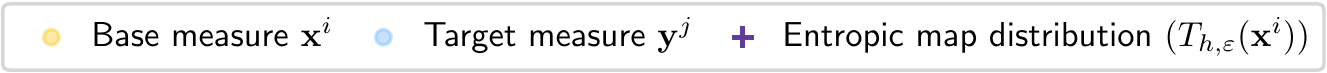}
    \end{subfigure}%
        \caption{Follow-up to \autoref{fig:entropic_map_illustration}, where a fresh sample of points from the base measure is transported using the various entropic map estimators $T_{h, \varepsilon}$ that were considered. Paired with \autoref{fig:entropic_map_illustration}, this plot shows the tradeoff, controlled by $\gamma$, between the sparsity of displacements and the ability to recover the target measure (as $\gamma$ increases, ultimately, the map no longer moves points.). An interesting feature of the $\ovk{\cdot}$ norm resides in its ability, no matter what $\gamma$, to enforce at least one displacement (here $k=1$).}
        \label{fig:entropic_move_illustration} 
    \end{figure*}

\section{Background}\label{sec:back}
\subsection{The Monge Problem.}
 Consider a translation-invariant cost function $c(\bx,\by):=h(\bx-\by)$, where $h:\R^d\rightarrow \R$. The \citeauthor{Monge1781} problem \citeyearpar{Monge1781} consists of finding, among all maps $T:\R^d\rightarrow \R^d$ that push-forward a measure $\mu\in\Pcal(\R^d)$ onto $\nu\in\Pcal(\R^d)$, the map which minimizes the average length (as measured by $h$) of its displacements:
 \begin{equation}\label{eq:monge}
     T^\star := \arginf_{T\sharp\mu = \nu} \int_{\R^d} h(\bx-T(\bx))\, \mathrm{d}\mu\,.
 \end{equation}

\paragraph{From Dual Potentials to Optimal Maps.} Problem~\eqref{eq:monge} is notoriously difficult to solve directly since the set of admissible maps $T$ is not even convex. The defining feature of OT theory to obtain an optimal push-forward solution $T^\star$ is to cast Problem~\eqref{eq:monge} as a linear optimization problem: relax the requirement that $\bx$ is mapped onto a single point $T(\bx)$, to optimize instead over the space of \textit{couplings} of $\mu, \nu$, namely on the set $\Pi(\mu,\nu)$ of probability distributions in $\Pcal(\R^d\times\R^d)$ with marginals $\mu,\nu$:
\begin{equation}\label{eq:primal}
     P^\star := \arginf_{P\in\Pi(\mu,\nu)} \iint_{\R^d\times\R^d} c\, \mathrm{d}P\,.
\end{equation}
If $T^\star$ is optimal for ~\autoref{eq:monge}, then $(\text{Id},T^\star)\sharp\mu$ is trivially an optimal coupling. To recover a map $T^\star$ from a coupling $P^\star$ requires considering the dual to \eqref{eq:primal}:
\begin{equation}\label{eq:dual}
     f^\star, g^\star \in \argsup_{\substack{f,g: \R^d\rightarrow \R\\f\oplus g\leq c}} \,\,\int_{\R^d} f\, \mathrm{d}\mu +\int_{\R^d} g \,\mathrm{d}\nu\,,
\end{equation}
where $\forall\, \bx,\by$ we write $(f\oplus g)(\bx,\by) := f(\bx)+ g(\by)$.

Leaving aside how such couplings and dual potentials can be approximated from data (this will be discussed in the next section), suppose that we have access to an optimal dual pair $(f^\star,g^\star)$.
By a standard duality argument~\citep[\S1.3]{santambrogio2015optimal}, if a pair $(\bx^0,\by^0)$ lies in the support of $P^\star$, $\supp(P^\star)$, the constraint for dual variables is saturated, i.e., 
$$f^\star(\bx^0) + g^\star(\by^0) = h(\bx^0 - \by^0)\,,$$
Additionally, by a so-called $c$-concavity argument one has: 
$$g^\star(\by^0)=\inf_{\bx} h(\bx - \by^0) - f^\star(\bx).\,$$ 
Assuming $f^\star$ is differentiable
at $\bx^0$, combining these two results yields perhaps the most pivotal result in OT theory: 
\begin{equation}\label{eq:prim-dual}
(\bx^0, \by^0) \in \supp(P^\star) \Leftrightarrow \nabla f^\star(\bx^0) \in \partial h(\bx^0-\by^0)\,,
\end{equation}
where $\partial h$ denotes the subdifferential of $h$, see, e.g.~\citep{carlier2010strategy}. Let $h^*$ be the \textit{convex conjugate} of $h$,
$$h^*(\by) := \sup_{x\in\R^d} \langle \bx,\by \rangle - h(\bx).$$
Depending on $h$, two cases arise in the literature:
\begin{itemize}[leftmargin=.3cm,itemsep=.0cm,topsep=0cm]
\item If $h$ is differentiable everywhere, strictly convex, one has:
$$\nabla f^\star(\bx^0) = \nabla h(\bx^0-\by^0)\,.$$ 
Thanks to the identity $\nabla h^* = (\nabla h)^{-1}$, one can uniquely characterize the only point $\by^0$ to which $\bx^0$ is associated in the optimal coupling $P^\star$ as  $\bx^0 - \nabla h^*(\nabla f^\star(\bx^0))$. More generally one recovers therefore for any $\bx$ in $\supp(\mu)$:
\begin{equation}\label{eq:brenier}
T^\star(\bx)=\bx - \nabla h^* \circ \nabla f^\star(\bx)\,.
\end{equation}
The \citeauthor{Bre91} theorem \citeyearpar{Bre91} is a particular case of that result, which states that when $h=\frac{1}{2}\|\cdot\|^2_2$, we have $T(\bx)=\bx - \nabla f^\star(\bx^0)$, since in that case $\nabla h=\nabla h^* = (\nabla h)^{-1}=\text{Id}$, see \citep[Theo.~1.22]{santambrogio2015optimal}.
\item If $h$ is ``only'' convex, then one recovers the sub-differential inclusion $\by^0 \in \bx^0 + \partial h^*(\nabla f^\star(\bx^0))$~\citep{ambrosio}\citep[\S3]{santambrogio2015optimal}.
\end{itemize}
In summary, given an optimal \textit{dual} solution $f^\star$ to Problem~\eqref{eq:dual}, one can use differential (or sub-differential) calculus to define an optimal transport map, in the sense that it defines (uniquely or as a multi-valued map) where the mass of a point $\bx$ should land.

\subsection{Bregman Centroids}
We suppose in this section that $h$ is strongly convex, in which case its convex conjugate is differentiable everywhere and gradient smooth. The generalized Bregman divergence (or B-function) generated by $h$~\citep{TelgarskyD12,kiwiel1997proximal} is, 
$$D_h(\bx|\by) = h(\bx) - h(\by) - \sup_{\bw\in\partial h(\by)} \langle \bw, \bx-\by\rangle.$$
Consider a family of $k$ points $\bz^1, \dots, \bz^m \in\R^d$ with weights $p^1,\dots, p^m >0$ summing to $1$. A point in the set
$$\argmin_{\bz\in\R^d} \sum_j p^{\,j} D_h(\bz,\bz^j)\,,$$ 
is called a Bregman centroid~\citep[Theo. 3.2]{nock}. Assuming $h$ is differentiable at each $\bz^j$, one has that this point is uniquely defined as:
\begin{equation}\label{eq:bregmanc}
\Ccal_h\left((\bz^j)_j, (p^{\,j})_i\right) := \nabla h^{\star}\left(\sum_{j=1}^m p^{\,j} \nabla h(\bz^i)\right)\,.
\end{equation}
\begin{figure}[t!]
       \centering
    \includegraphics[width=.47\columnwidth]{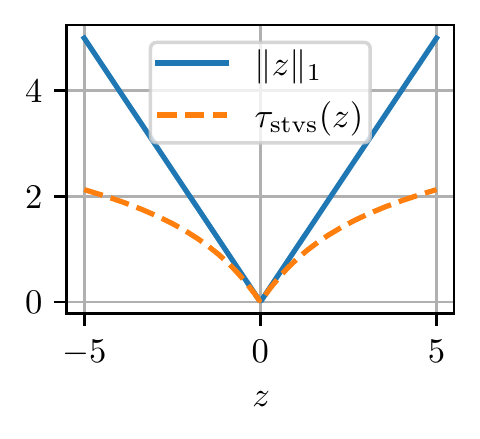}
    \includegraphics[width=.5\columnwidth]{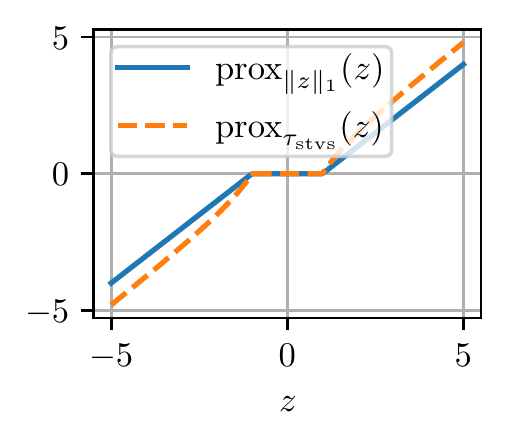}
    \vspace{-1.5em}
        \caption{The STVS regularizer is not convex, but its proximal operator is well-defined and tends to shrink values less than the usual soft-thresholding operator. For instance, its values near $\{-5,5\}$ are close to the identity line.}
        \label{fig:prox} 
\end{figure}

\subsection{Sparsity-Inducing Penalties}
To form relevant functions $h$, we will exploit the following sparsity-inducing functions: the $\ell_1$ and $\ovk{\cdot}$ norms, and a handcrafted penalty that mimics the thresholding properties of $\ell_1$ but with less shrinkage.
\begin{itemize}[leftmargin=.3cm,itemsep=.0cm,topsep=0cm]
\item For a vector $\bz\in\R^d$, $\|\bz\|_p:=(\sum_{i=1}^d |\bz_i|^p)^{1/p}$. We write $\ell_2$ and $\ell_2^2$ for $\|\cdot\|_2$ and $\|\cdot\|_2^2$ respectively.
\item We write $\ell_1$ for $\|\cdot\|_1$. Its proximal operator $\mathrm{prox}_{\gamma \ell_1}(\bz) = \mathrm{ST}_\gamma(\bz) = (1 - \gamma/|\bz|)_+\odot \bz$ is called the soft-thresholding operator.
%
%
\item\citet{schreck2015shrinkage} propose the soft-thresholding operator with vanishing shrinkage (STVS),
\begin{equation}\label{eq:svts}
\tau_{\mathrm{stvs}}(\bz) = \gamma^2\mathbf{1}_d^T\left(\sigma(
\bz)+\tfrac{1}{2}-\tfrac{1}{2}e^{-2\sigma(
\bz)}\right) \geq 0\,,
\end{equation}
with $\sigma(
\bz):= \arcsinh\left(\frac{\bz}{2\gamma}\right)$, 
and where all operations are element-wise. $\tau_{\mathrm{stvs}}$ is a non-convex regularizer, non-negative thanks to (our) addition of $+\tfrac{1}{2}$, to recover a nonnegative quantity that cancels if and only if $\bz=0$.~\citeauthor{schreck2015shrinkage} show that the proximity operator $\prox_{\tau_{\mathrm{stvs}}}$, written $\mathrm{STVS}$ for short, decreases the shrinkage (see ~\autoref{fig:prox}) observed with soft-thresholding:
\begin{equation}\label{eq:proxstvs}
\mathrm{STVS}_\gamma(\bz) = \left(1- \gamma^2/|\bz|^2\right)_+ \odot \bz\,.
\end{equation}
The Hessian of $\tau_{\mathrm{stvs}}$ is a diagonal matrix with values $\tfrac{1}{2} |\bz| / \sqrt{\bz^2 + \gamma ^2} - \tfrac{1}{2}$ and is therefore lower-bounded (with positive-definite order) by $-\tfrac{1}{2}I_d$.
\item Let $\mathcal{G}_k$ be the set of all subsets of size $k$ within $\{1,\dots, d\}$. \citet{argyriou} introduces the $k$-overlap norm:
$$
\ovk{\bz} = \min\{\! \sum_{I\in\mathcal{G}_k} \|\bv_I\|_2\, |\, \text{supp}(\bv_I)\subset I, \sum_{I\in\mathcal{G}_k} \bv_I = \bz\}\,.
$$
For any vector $\bz$ in $\R^d$, we write $\bz^\downarrow$ for the vector composed with all entries of $\bz$ sorted in a \textit{decreasing} order. This formula can be evaluated as follows to exhibit a $\ell_1/\ell_2$ norm split between the $d$ variables in a vector:
$$
\ovk{\bz}^2 = \sum_{i=1}^{k-r-1} (|\bz|^\downarrow_i)^2 + \left(\sum_{i=k-r}^d |\bz|^\downarrow_i\right)^2/(r+1)
$$
where $r\leq k-1$ is the unique integer such that  $$|\bz|_{k-r}^\downarrow \leq \sum_{i=k-r}^d |\bz|^\downarrow_i <|\bz|_{k-r-1}^\downarrow.$$
Its proximal operator is too complex to be recalled here but given in~\citep[Algo. 1]{argyriou}, running in $O(d(\log d+k))$ operations.
\end{itemize}%
Note that both $\ell_1$ and $\tau_{\mathrm{stvs}}$ are separable---their proximal operators act element-wise---but it is not the case of $\|\cdot\|_{\mathrm{ov}k}$.

\section{Generalized Entropic-Bregman Maps}\label{sec:methods}

\textbf{Generalized Entropic Potential. } 
When $h$ is the $\ell_2^2$ cost, and when $\mu$ and $\nu$ can be accessed through samples, i.e., $\hat\mu_n = \tfrac{1}{n}\sum_i \delta_{\bx^i},\hat\nu_m=\tfrac{1}{m}\sum_j \delta_{\by^j}$, a convenient estimator for $f^\star$ and subsequently $T^\star$ is the entropic map~\citep{pooladian2021entropic,rigollet2022sample}.  We generalize these estimators for arbitrary costs $h$. Similar to the original approach, our construction starts by solving a dual entropy-regularized OT problem. Let $\varepsilon>0$ and write $K_{ij} = [\exp(-h(\bx^i-\by^j)/\varepsilon)]_{ij}$ the kernel matrix induced by cost $h$. Define (up to a constant):
\begin{equation}\label{eq:finitedual}
\!\!\!\bff^\star, \bg^\star = \argmax_{\bff\in\R^n,\bg\in\R^m} \langle\bff, \tfrac{\mathbf{1}_n}{n}\rangle + \langle\bg, \tfrac{\mathbf{1}_m}{m}\rangle  - \varepsilon \langle e^{\frac{\bff}{\varepsilon}}, K e^{\frac{\bg}{\varepsilon}}\rangle\,.
\end{equation}
Problem~\eqref{eq:finitedual} is the regularized OT problem in dual form~\citep[Prop.~4.4]{Peyre2019computational}, an unconstrained concave optimization problem that can be solved with the \citeauthor{Sinkhorn64} algorithm~\citep{cuturi2013sinkhorn}. Once such optimal vectors are computed, estimators $f_\varepsilon, g_\varepsilon$ of the optimal dual functions $f^\star, g^\star$ of \autoref{eq:dual} can be recovered by extending these discrete solutions to unseen points $\bx, \by$,
\begin{align}\label{eq:fdual}
f_\varepsilon(\bx)= \lse_\varepsilon([h(\bx-\by^j) - \bg^\star_j]_j)\,,\\
\label{eq:gdual}
g_\varepsilon(\by)= \lse_\varepsilon([h(\bx^i-\by) - \bff^\star_i]_i)\,,
\end{align}
where for a vector $\bu$ or arbitrary size $s$ we define the log-sum-exp operator as $\lse_\varepsilon(\bu):= - \varepsilon \log (\tfrac{1}{s}\mathbf{1}_s^Te^{-\bu/\varepsilon})$.

\paragraph{Generalized Entropic Maps.} Using the blueprint given in~\autoref{eq:prim-dual}, we use the gradient of these dual potential estimates to formulate maps. Such maps are only properly defined on a subset of $\R^d$ defined as follows: 
\begin{equation}
\Omega_{\hat{\nu}_m}(h) := \{ \bx \,|\, \forall j \leq m, \nabla h (\bx - \by^j) \text{ exists.} \} \subset\R^d.
\end{equation}
However, because a convex function is a.e. differentiable, $\Omega_{\hat{\nu}_m}(h)$ has measure 1 in $\R^d$. With this, $\nabla f_\varepsilon$ is properly defined for $\bx$ in $\Omega_{\hat{\nu}_m}(h)$, as:
\begin{align}\label{eq:gradfeps}
\nabla f_\varepsilon(\bx)= \sum_{j=1}^m p^{\,j}(\bx) \nabla h(\bx-\by^j)\,,
\end{align}
using the $\bx$-varying Gibbs distribution in the $m$-simplex:
\begin{equation}\label{eq:gibbs}
p^{\,j}(\bx):=\frac{\exp\left(- \left(h(\bx-\by^j)-\bg_j^\star\right)/\varepsilon\right)}{\sum_{k=1}^m \exp\left(- \left(h(\bx-\by^k)-\bg_k^\star\right)/\varepsilon\right)}\,.
\end{equation} 
One can check that if $h=\tfrac{1}{2}\ell^2_2$, \autoref{eq:gradfeps} simplifies to the usual estimator~\citep{pooladian2021entropic}: 
\begin{equation}
T_{2,\varepsilon}(\bx) : = \bx - \nabla f_\varepsilon(\bx) = \sum_{j=1}^m p^{\,j}(\bx)\by^j\,.
\end{equation}
We can now introduce the main object of interest of this paper, starting back from~\autoref{eq:brenier}, to provide a suitable generalization for entropic maps of elastic-type:
\begin{definition}\label{def:emap} The entropic map estimator for $h$ evaluated at $\bx\in\Omega_{\hat{\nu}_m}(h)$ is $\bx-\nabla h^*\circ\nabla f_\varepsilon(\bx)$. This simplifies to:
\begin{equation}\label{eq:T-estimator}T_{h,\varepsilon}(x) := \bx - \Ccal_h((\bx-\by^j)_j, (p^{\,j}(\bx))_j)\end{equation}
\end{definition}

\begin{figure}[t!]
\centering
\includegraphics[width=0.49\textwidth]{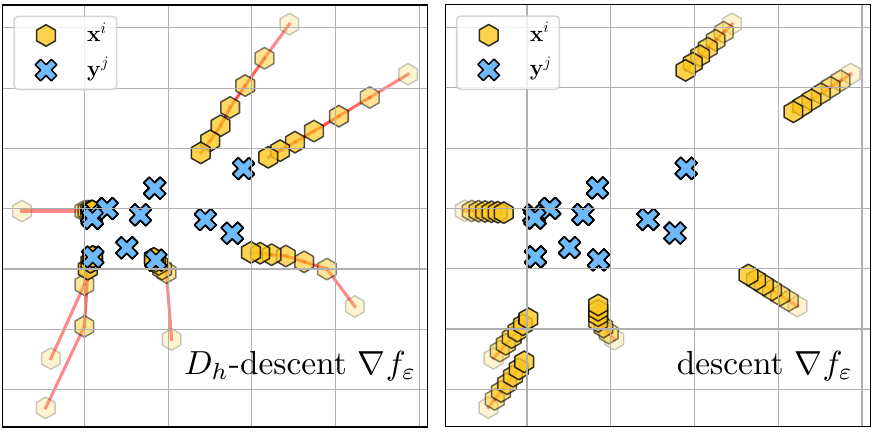}
\caption{Difference between $W_c$ gradient descents, minimizing loss directly in direction $-\lambda\nabla f_\varepsilon$, or Bregman descent as described in \autoref{eq:dhdescent} when $h=\tfrac12\ell_2^2 +\ell_1$, see \autoref{prop:displ} for details on computing $\nabla h^*$. Six steps are plotted with stepsize $\lambda=\tfrac14$.} \label{fig:differentgrads}
\end{figure}

\paragraph{Bregman Centroids \textit{vs.} $W_c$ Gradient flow}
To displace points, a simple approach consists of following $W_c$ gradient flows, as proposed, for instance, in~\citep{pmlr-v32-cuturi14} using a primal formulation~\autoref{eq:primal}. In practice, this can also be implemented by relying on variations in dual potentials $\nabla f_\varepsilon$, as advocated in~\citet[\S4]{feydy2019interpolating}. This approach arises from the approximation of $W_c(\hat\mu_n,\hat\nu_m)$ using the dual objective~\autoref{eq:dual},
$$
S_{h,\varepsilon} \left(\tfrac{1}{n}\textstyle\sum_i \delta_{\bx_i}, \tfrac{1}{m}\textstyle\sum_j \delta_{\by_j}\right) = \tfrac{1}{n}\textstyle\sum_i f_\varepsilon(\bx_i) + \tfrac{1}{m}\textstyle\sum_j g_\varepsilon(\by_j)\,,
$$
differentiated using the Danskin theorem. As a result, any point $\bx$ in $\mu$ is then pushed away from $\nabla f_\varepsilon$ to decrease that distance. This translates to a gradient descent scheme:
$$
\bx \leftarrow \bx - \lambda\nabla f_\varepsilon(\bx)
$$
Our analysis suggests that the descent must happen relative to $D_h$, to use, instead, a Bregman update (here $\bar\lambda=1-\lambda$):
\begin{equation}\label{eq:dhdescent}
\bx \leftarrow \nabla h^*\!\!\left( \bar\lambda \nabla h (\bx) + \lambda \nabla h(\bx - \nabla h^*\!\!\circ\nabla f_\varepsilon(\bx))\right)
\end{equation}
Naturally, these two approaches are exactly equivalent as $h=\tfrac{1}{2}\ell_2^2$ but result in very different trajectories for other functions $h$ as shown in~\autoref{fig:differentgrads}.
\section{Structured Monge Displacements}\label{sec:methods2}
We introduce in this section cost functions $h$ that we call of elastic-type, namely functions with a $\ell_2^2$ term in addition to another function $\tau$. When $\tau$ is sparsity-inducing (minimized on sparse vectors, with kinks) and has a proximal operator in closed form, we show that the displacements induced by this function $h$ are feature-sparse.

\subsection{Elastic-type Costs}
By reference to~\citep{zou2005regularization}, we call $h$ of elastic-type if it is strongly convex and can be written as 
\begin{equation}
\label{eq:reg_l2}
    h(\bz) := \tfrac{1}{2}\|\bz\|^2 + \tau(\bz)\,.
\end{equation}
where $\tau:\R^d \rightarrow \R$ is a function whose proximal operator is well-defined. Since OT algorithms are invariant to a positive rescaling of the cost $c$, our elastic-type costs subsume, without loss of generality, all strongly-convex translation invariant costs with convex $\tau$. They do also include useful cases arising when $\tau$ is not (e.g., $\tau_{\mathrm{stvs}}$).
\begin{proposition}\label{prop:displ} For $h$ as in \eqref{eq:reg_l2} and $\bx\in\Omega_{\hat\nu_m}(\tau)$ one has:
\begin{equation}\label{eq:T-estimator-prox}T_{h,\varepsilon}(x) := \bx - \prox_\tau\!\!\left(\!\bx - \sum_{j=1}^m p^{\,j}(\bx) \left(\by^j + \nabla \tau(\bx-\by^j)\right)\!\!\right)\end{equation}
\end{proposition}
\begin{proof}
The result follows from $\nabla h^* = \prox_{\tau}.$ Indeed:
\begin{align*}
h^*(\bw) &= \sup_{\bz} \bw^T \bz - \tfrac{1}{2}\|\bz\|^2 - \tau(\bz)\\
&= - \inf_{\bz} - \bw^T \bz + \tfrac{1}{2} \|\bz\|^2 + \tau(\bz)\\
&= \tfrac{1}{2}\|\bw\|^2 - \inf_\bz \tfrac{1}{2}\|\bz-\bw\|^2 + \tau(\bz).
\end{align*}
Differentiating on both sides and using Danskin's lemma, we get the desired result by developing $\nabla h$ and taking advantage of the fact that the weights $p^{\,j}(\mathbf{x})$ sum to 1.
\end{proof}

\subsection{Sparsity-Inducing Functions $\tau$}
We discuss in this section the three choices we introduced in \S\ref{sec:back} for proximal operators and their practical implications in the context of our generalized entropic maps.
\paragraph{1-Norm $\ell_1$.}
As a first example, we consider $\tau(\bz) = \gamma \|\bz\|_1$ in \autoref{eq:reg_l2}.
The associated proximal operator is the soft-thresholding operator $\mathrm{prox}_{\tau}(\cdot) = \operatorname{ST}(\cdot,\gamma)$ mentioned in the introduction.
We also have $\nabla h (\bz) = \bz + \gamma\, \sign(\bz)$ for $\bz$ with no $0$ coordinate.
Plugging this in \autoref{eq:brenier}, we find that the Monge map $T_{\gamma \ell_1,\varepsilon}(\bx)$ is equal to
$$
 \bx - \mathrm{ST}_\gamma\left( \bx - \sum_{j=1}^m p^{\,j}(\bx) \left(\by^j + \gamma \sign(\bx - \by^j)\right)\right)\,,
$$
where the $p^{\,j}(\bx)$ are evaluated at $\bx$ using \autoref{eq:gibbs}.
Applying the transport consists in an element-wise operation on $\bx$: for each of its features $t\leq d$, one substracts $\mathrm{ST}_\gamma\left( \sum_{j=1}^m p^{\,j}(\bx)\nabla h(\bx_t - \by^j_t)\right)$.
The only interaction between coordinates comes from the weights $p^{\,j}(\bx)$.

The soft-thresholding operator sparsifies the displacement. Indeed, when for a given $\bx$ and a feature $t\leq d$ one has
$$|\bx_t - \sum_{j=1}^mp^{\,j}(\bx)\by^j_t + \gamma\sign(\bx_t-\by^j_t)| \leq \gamma,$$ then there is no change on that feature : $[T_{\gamma \ell_1,\varepsilon}(\bx)]_t = \bx_t$.
That mechanism works to produce, locally, sparse displacements on certain coordinates. Another interesting phenomenon happens when $\bx$ is too far from the $\by_j$'s on some coordinates, in which case the transport defaults back to a $\ell_2^2$ average of the target points $\by^j$ (with weights that are, however, influenced by the $\gamma\ell_1$ regularization):
\begin{proposition}\label{prop:testimator}
    If $\bx$ is such that $\bx_t\geq \max_j \by^{j}_{t}$ or $\bx_t\leq \min_j \by^j_t$ then $T_{\gamma \ell_1,\varepsilon}(\bx)_t = \sum_j p^{\,j}(\bx)\by^j_t$.
\end{proposition}
\begin{proof}
For instance, assume $\bx_t\geq \max_j \by^j_t$.
Then, for all $j$, we have $\mathrm{sign}(\bx_t - \by^j_t) = 1$, and as a consequence $\sum_{j=1}^mp^{\,j}(\bx)\nabla h(\bx - \by^j)_t = \bx_t - \sum_jp^{\,j}(\bx)\by^j_t + \gamma$.
This quantity is greater than $\gamma$, so applying the soft-thresholding gives $\mathrm{ST}_\gamma(\sum_{j=1}^mp^{\,j}(\bx)\nabla h(\bx - \by^j)_t) = \bx_t - \sum_jp^{\,j}(\bx)\by^j_t$,
which gives the advertised result. Similar reasoning gives the same result when $\bx_t\leq \min_j \by^j_t$.
\end{proof}
Interestingly, this property depends on $\gamma$ only through the $p^{\,j}(\bx)$'s, and the condition that  $\bx_t\geq \max_j \by^{j}_{t}$ or $\bx_t\leq \min_j \by^j_t$ does not depend on $\gamma$ at all.

\paragraph{Vanishing Shrinkage STVS:} The $\ell_1$ term added to form the elastic net has a well-documented drawback, notably for regression: on top of having a sparsifying effect on the displacement, it also shrinks values. This is clear from the soft-thresholding formula, where a coordinate greater than $\gamma$ is reduced by $\gamma$. This effect can lead to some ``shortening'' of displacement lengths in the entropic maps.
We use the Soft-Thresholding with Vanishing Shrinkage (STVS) proposed by \citet{schreck2015shrinkage} to overcome this problem.
The cost function is given by \autoref{eq:svts}, and its prox in~\autoref{eq:proxstvs}. When $|\bz|$ is large, we have $\mathrm{prox}_{\tau_{\mathrm{stvs}}}(\bz) = \bz + o(1)$, which means that the shrinkage indeed vanishes.
Interestingly, even though the cost $\tau_{\mathrm{stvs}}$ is non-convex, it still has a proximal operator, and $\frac12\|\cdot\|^2 + \tau_{\mathrm{stvs}}$ is $\tfrac12$-strongly convex.

\paragraph{$k$-Overlap: $\tau=\ovk{\cdot}$.} The $k$-overlap norm offers the distinctive feature that its proximal operator 
selects anywhere between $d$ (small $\gamma$) and $k$ (large $\gamma$) non-zero variables, see~\autoref{fig:entropic_map_illustration}. Applying this proximal operator is, however, significantly more complex, because it is not separable across coordinates and requires $d(k+\log d)$ operations, instantiating a $k\times d-k$ matrix to select two integers $r,l$ ($r\leq k \leq l$) at each evaluation. We were able to use it for moderate problem sizes but these costs became prohibitive on larger scale datasets, where $d$ is a few tens of thousands.
\section{Experiments}
\label{sec-experiments-main}
We start this experimental study with two synthetic tasks.
For classic costs such as $\ell_2^2$, several examples of ground-truth optimal maps are known. Unfortunately, we do not know yet how to propose ground-truth $h$-optimal maps, nor dual potentials, when $h$ has the general structure considered in this work.
As a result, we study two synthetic problems, where the sparsity pattern of a ground truth transport is either constant across $\bx$ or split across two areas.
We follow with an application to single-cell genomics, where the modeling assumption that a treatment has a sparse effect on gene activation (across $34k$ genes) is plausible. 
In terms of implementaiton, the entire pipeline described in \S~\ref{sec:methods} and \ref{sec:methods2} rests on running the Sinkhorn algorithm first, with an appropriate cost, and than differentiating the resulting potentials. This can be carried out in a few lines of code using a parameterized \texttt{TICost}, fed into the \texttt{Sinkhorn} solver, to output a \texttt{DualPotentials} object in \textsc{OTT-JAX}\footnote{\texttt{https://github.com/ott-jax/ott}}\citep{cuturi2022optimal}. We a class of regularized translation invariant cost functions, specifying both regularizers $\tau$ and their proximal operators. We call such costs \texttt{RegTICost}.

\subsection{Synthetic experiments.}

\begin{figure}[t!]
       \centering
        \includegraphics[width=\columnwidth]{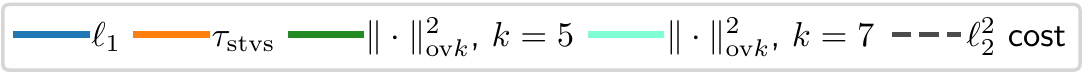}\\
    \includegraphics[width=.33\columnwidth]{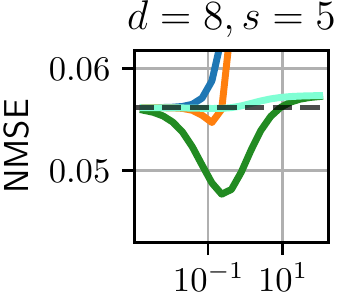}
    \hspace{.02\columnwidth}
    \includegraphics[width=.295\columnwidth]{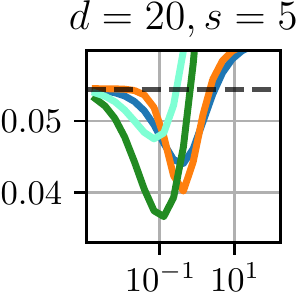}
    \includegraphics[width=.325\columnwidth]{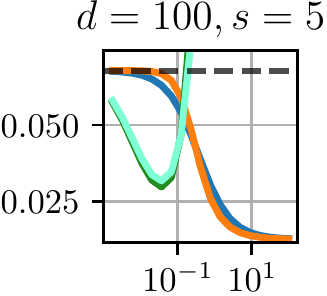}\\

        \includegraphics[width=0.358\columnwidth]{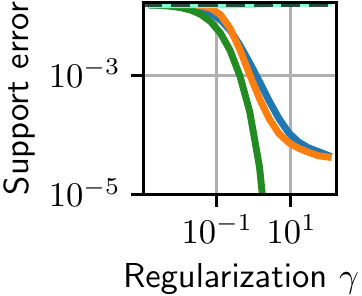}
    \includegraphics[width=0.311\columnwidth]{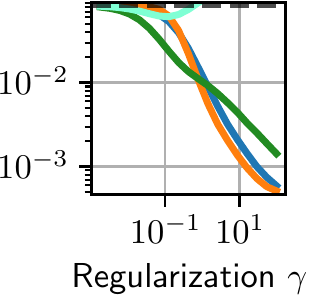}
    \includegraphics[width=0.311\columnwidth]{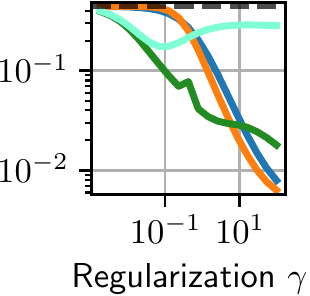}\\
        \caption{\textbf{Synthetic experiment:} ability of the estimated $T_{h,\varepsilon}$ to recover a ground-truth transport which displaces $s$ coordinates in dimension $d$, with $n=1000$ samples. We compare the different costs $h$ proposed in the paper with the classical $\ell_2^2$ cost. We identify $3$ regimes. \textbf{Left:} when $s\simeq d$, the $\ell_2^2$ cost is already good, and the proposed costs barely improve over it in terms of MSE. \textbf{Middle:} when $d$ is moderately larger than $s$, all the proposed costs improve over the $\ell_2^2$ cost, and the optimal regularization for $\ell_1$ and $\tau_{\mathrm{stvs}}$ are finite. 
        \textbf{Right:} When $d \gg s$, the proposed methods vastly improve over the $\ell_2^2$ cost. The optimal regularization for $\ell_1$ and $\tau_{\mathrm{stvs}}$ is infinite even for the MSE. In terms of support error, larger regularization always leads to better results.}
        \label{fig:synth_expe} 
\end{figure}

\textbf{Constant sparsity-pattern. } We measure the ability of our method to recover a sparse transport map using
a setting inspired by \citep{pooladian2021entropic}.
Here $\mu=\mathcal{U}_{[0, 1]^d}$. For an integer $s < d$, we set $\nu= T^\star_s\sharp\mu$, where the map $T^\star_s$ acts on coordinates independently with the formula $T^\star_s(\bx) = [\exp(\bx_1), \dots, \exp(\bx_s), \bx_{s+1},\dots, \bx_d]$: it only changes the first $s$ coordinates of the vector, and corresponds to a sparse displacement when $s \ll d$.
Note that this sparse transport plan is much simpler than the maps our model can handle since, for this synthetic example, the sparsity pattern is fixed across samples. Note also that while it might be possible to detect that only the first $s$ components have high variability using a 2-step pre-processing approach, or an adaptive, robust transport approach~\citep{paty2019subspace}, our goal is to detect that support in a one-shot, thanks to our choice of $h$. 
We generate $n=1,000$ i.i.d. samples $\bx^i$ from $\mu$, and $\by^j$ from $\nu$ independently; the samples $\by^j$ are obtained by first generating fresh i.i.d. samples $\tilde{\bx}^j$ from $\mu$ and then pushing them : $\by^j := T^\star_s(\tilde{\bx}^j)$. We use our three costs to compute $T_{h, \varepsilon}$ from these samples, and measure our ability to recover $T^\star_s$ from $T_{h, \varepsilon}$ using a normalized MSE defined as $\frac1{nd}\sum_{i=1}^n \|T^\star_s(\bx^i) - T_{h, \varepsilon}(\bx^i)\|^2$.
We also measure how well our method identifies the correct support: for each sample, we compute the \emph{support error} as $\sum_{i=s+1}^d\mathbf{\Delta}_i^2 /  \sum_{i=1}^d\mathbf{\Delta}_i^2$ with $\mathbf{\Delta}$ the displacement $T_{h, \varepsilon}(\bx)-\bx$. This quantity is between $0$ and $1$ and cancels if and only if the displacement happens only on the correct coordinates. We then average this quantity overall the $\bx^i$.
\autoref{fig:synth_expe} displays the results as $d$ varies and $s$ is fixed.
Here, $\tau_{\mathrm{stvs}}$ performs better than $\ell_1$.

\begin{figure}[t!]
\centering
  \begin{minipage}[c]{0.6\columnwidth}
    \includegraphics[width=\textwidth]{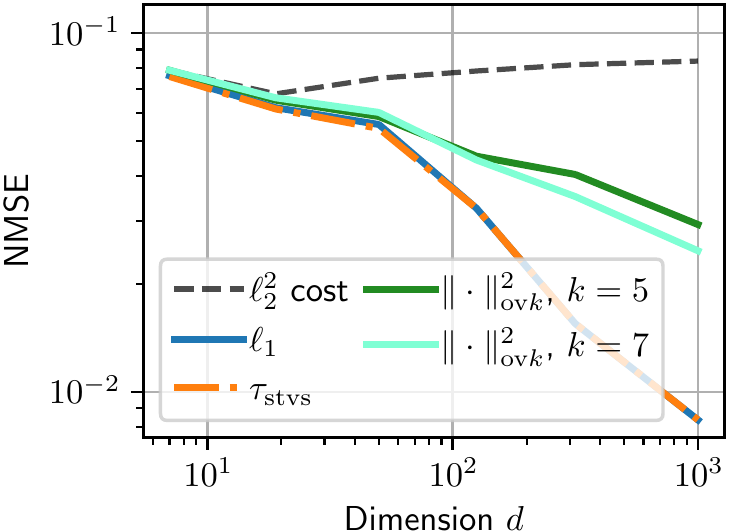}
  \end{minipage}\hfill
  \begin{minipage}[c]{0.38\columnwidth}
    \caption{\textbf{Scaling with dimension:} The number of samples is fixed to $n=100$, and the sparsity to $s=2$. For each dimension, we do a grid search over $\gamma$ and retain the one with the lowest MSE.
    } \label{fig:synth2}
  \end{minipage}
\end{figure}

\textbf{$\bx$-dependent sparsity-pattern. } To illustrate the ability of our method to recover transport maps whose sparsity pattern is adaptive, depending depending on the input $\bx$, we extend the previous setting as follows.
To compute $F_s(\bx)$, we compute first the norms of two coordinate groups of $\bx$: $n_1= \sum_{i=1}^s \bx_i^2$ and $n_2= \sum_{i=s+1}^{2s} \bx_i^2$. Second, we displace the coordinate group with the largest norm: if $n_1> n_2$, $F_s(\bx) = [\exp(\bx_1), \dots, \exp(\bx_s), \bx_{s+1},\dots, \bx_d]$, otherwise $F_s(\bx) = [\bx_1, \dots, \bx_s, \exp(\bx_{s+1}),\dots, \exp(\bx_{2s}), \bx_{2s+1}, \dots, \bx_d]$. Obviously, the displacement pattern depends on $\bx$.
\autoref{fig:synth2} shows the NMSE with different costs when the dimension $d$ increases while $s$ and $n$ are fixed.
As expected, we observe a much better scaling for our costs than for the standard $\ell^2_2$ cost, indicating that sparsity-inducing costs mitigate the curse of dimensionality.

\begin{figure*}[t!]
   \centering
\includegraphics[width=.245\textwidth]{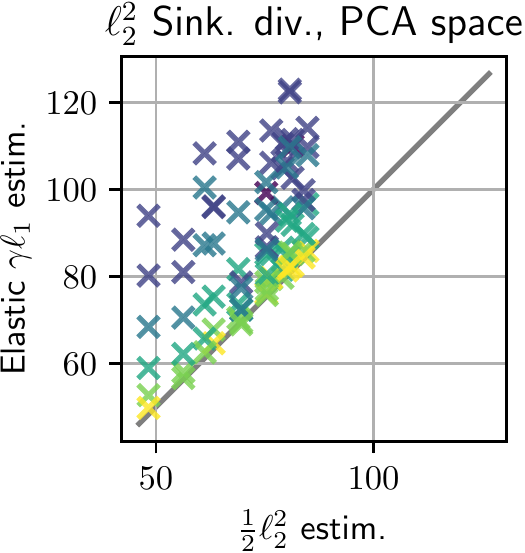}
\includegraphics[width=.22\textwidth]{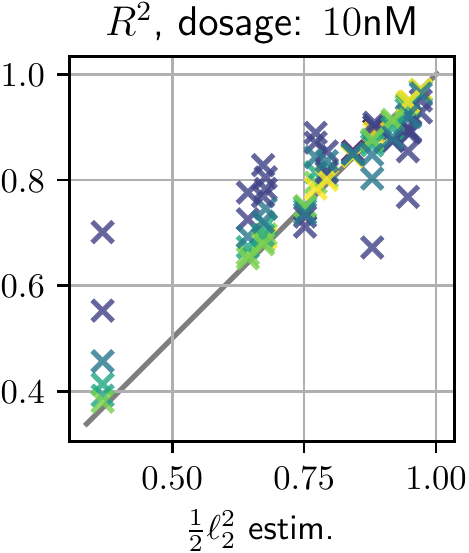}
\includegraphics[width=.218\textwidth]{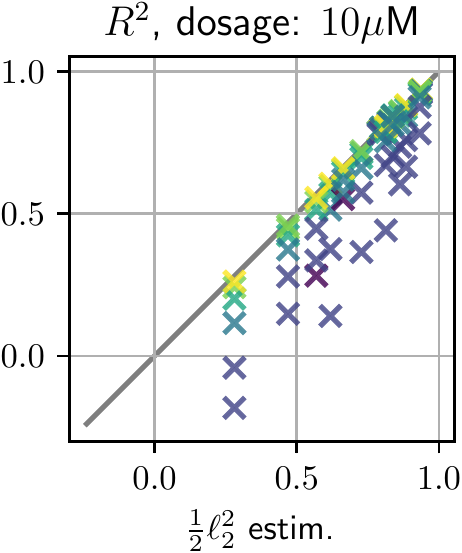}
\includegraphics[width=.215\textwidth]{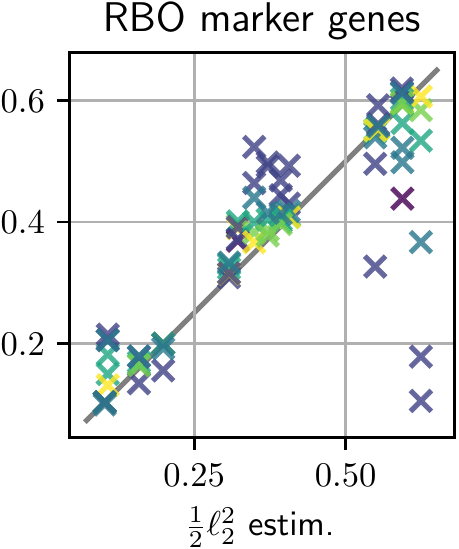}
\includegraphics[width=.055\textwidth]{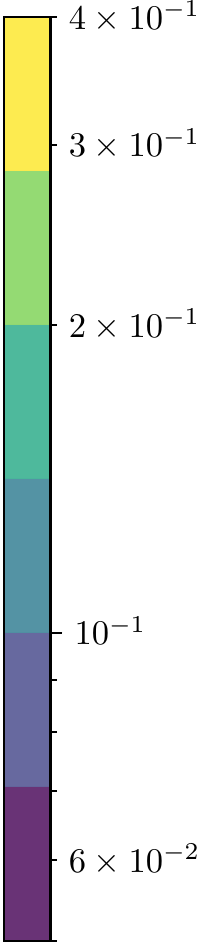}\\
\vspace{1em}

\includegraphics[width=.21\textwidth]{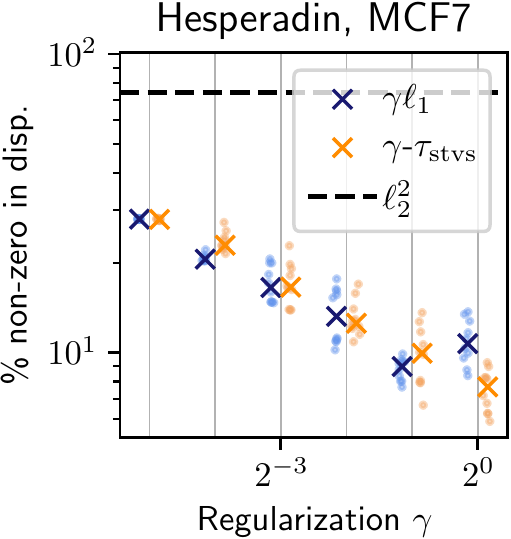}
\includegraphics[width=.22\textwidth]{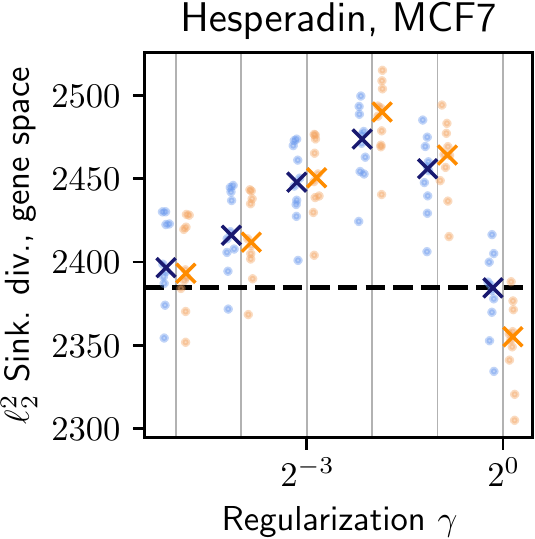}
\includegraphics[width=.21\textwidth]{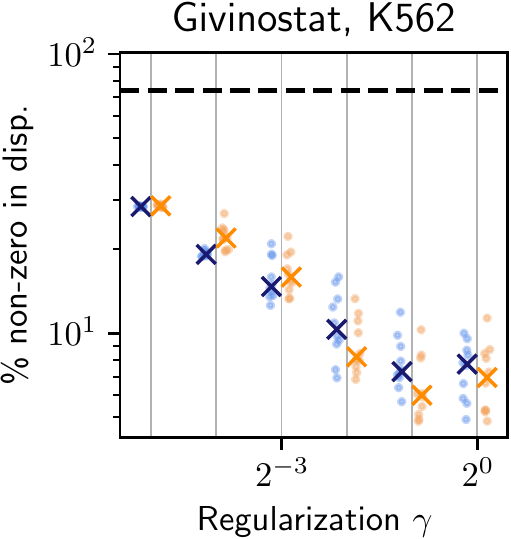}
\includegraphics[width=.22\textwidth]{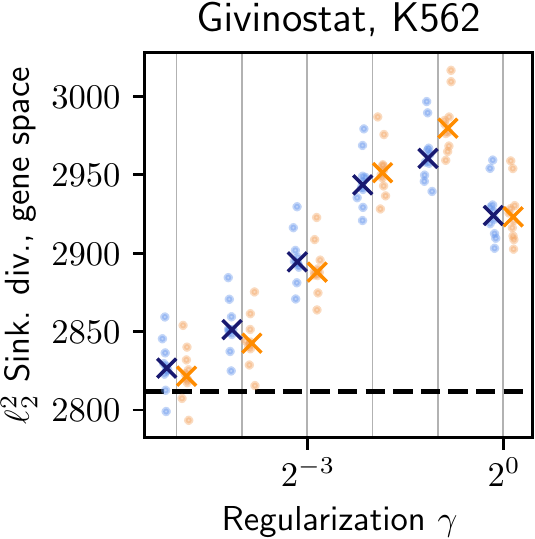}
    \caption{\textit{Top row}: performance, for all 15 experiments, of the elastic-$\gamma \ell_1$ estimator \textit{vs.} the  $\tfrac12\ell_2^2$ entropic map. We consider 6 values for $\gamma$. Each of the $15\times6$ crosses denotes the mean, over 10 random 80\%,/20\% splits of that cell line/drug experiment, of a quantity of interest. To facilitate reading, rather than reporting the $\gamma$ value, we report the average percentage of non-zero displacements (using \texttt{np.isclose(.,0)}, equivalently thresholding values below $10^{-8}$)  across all displaced points in that fold (yellow means $40\%$ dense displacements, dark-blue displacements only happen on $\approx 5\%$ of genes). While all these maps are estimated in full genes space ($\approx 34k$), we provide a simplified measure of their ability to reconstruct the measures by computing a $\tfrac12\ell_2^2$-Sinkhorn divergence in PCA space. This picture shows that one can sparsify significantly $\tfrac12\ell_2^2$ maps and still get a similar reconstruction error. Next, we picture separately the $R^2$ (see text body for details) computed on marker genes on low ($10nM$) and high ($10\mu M$) dosages of the drug. For low dosages, inducing sparsity in displacements seems to help, whereas this may no longer be the case when the effect of perturbations becomes large. Finally, the RBO metric shows that sparsity does help to select marker genes based only on map estimation. \textit{Bottom row}: Close up on Hesperadin/MCF7 and Givinostat/K562 experiments. For each, we quantify the sparsifying effect w.r.t $\gamma$, as well as $\tfrac12\ell_2^2$-Sinkhorn divergence in full gene space.}
    \label{fig:real_data} 
\end{figure*}

\subsection{Single-Cell RNA-seq data.}
We validate our approach on the single-cell RNA sequencing perturbation data from \citep{srivastan:19}. After removing cells with less than $200$ expressed genes and genes expressed in less than $20$ cells, the data consists of $579,483$ cells and $34,636$ genes. In addition, the raw counts have been normalized and $\log(x + 1)$ scaled.
We select the $5$ drugs (Belinostat, Dacinostat, Givinostat, Hesperadin, and Quisinostat) out of $188$ drug perturbations that are highlighted in the original data \citep{srivastan:19} as showing a strong effect.
We consider $3$ human cancer cell lines (A549, K562, MCF7) to each of which is applied each of the $5$ drugs. We use our four methods to learn an OT map from control to perturbed cells in each of these $3\times 5$ scenarios. For each cell line/drug pair, we split the data into $10$ non-overlapping 80\%/20\% train/test splits, keeping the test fold to produce our metrics.

\begin{table}
\centering
\caption{Per cell line, sample sizes of control + drug perturbation.}
\label{tab:sc-size}
\small{\begin{tabular}{lrrrrrr}
\toprule
{} &  Cont. &  Dac. &  Giv. &  Bel. &  Hes. &  Quis. \\
\midrule
\textbf{A549} &     3274 &         558 &         703 &         669 &         436 &          475 \\
\textbf{K562} &     3346 &         388 &         589 &         656 &         624 &          339 \\
\textbf{MCF7} &     6346 &        1562 &        1805 &        1684 &         882 &         1520 \\
\bottomrule
\end{tabular}}
\end{table}

\textbf{Methods.}
We ran experiments in two settings, using the whole $34,000$-$d$ gene space and subsetting to the top $5k$ highly variable genes using \textsc{ScanPy}~\citep{wolf2018scanpy}. We consider entropic map estimators with the following cost functions and pre-processing approaches:
$\tfrac12\ell_2^2$ cost; the $\tfrac12\ell_2^2$ cost on $50$-$d$ PCA space (PCA directions are recomputed on each train fold); Elastic with $\gamma\ell_1$; Elastic with $\gamma$-$\tau_{\mathrm{svts}}$ cost. We vary $\gamma$ for these two methods. We did not use the $\ovk{\cdot}$ norm because of memory challenges when handling such a high-dimensional dataset.  For the non-PCA-based approaches, we can also measure their performance in PCA space by projecting their high-dimensional predictions onto the 50-$d$ space. The $\varepsilon$ regularization parameter for all these approaches is set for each cost and experiment to $10\%$ of the mean value of the cost matrix between the train folds of control and treated cells, respectively.

\textbf{Evaluation.}
We evaluate methods using these metrics: 
\begin{itemize}[leftmargin=.3cm,itemsep=.0cm,topsep=0cm,parsep=2pt]
\item the $\ell_2^2$-Sinkhorn divergence (using $\varepsilon$ to be 10\% of the mean of pairwise $\ell_2^2$ cost matrix of treated cells) between transferred points (from test fold of control) and test points (from perturbed state); \textit{lower is better}.
\item Ranked biased overlap~\citep{webber2010similarity} with $p=0.9$, between the $50$ perturbation marker genes as computed on all data with \textsc{ScanPy}, and the following per-gene statistic, computed using a map as follows: average (on fold) expression of (predicted) perturbed cells from original control cells (this tracks changes in log-expression before/after predicted treatment); \textit{higher is better}.
\item Coefficient of determination ($R^2$) between the average ground-truth / predicted gene expression on the $50$ perturbation markers~\citep{Lotfollahi2019}; \textit{higher is better}.
\end{itemize}
These results are summarized in \autoref{fig:real_data}, across various costs, perturbations and hyperparameter choices.

\textbf{Conclusion.} We consider structured translation-invariant ground costs $h$ for transport problems. After forming an entropic potential with such costs, we plugged it in~\citeauthor{Bre91}'s approach to construct a generalized entropic map. We highlighted a surprising connection between that map and the Bregman centroids associated with the divergence generated by $h$, resulting in a more natural approach to gradient flows defined by $W_c$, illustrated in a simple example. By selecting costs $h$ of elastic type (a sum of $\ell_2^2$ and a sparsifying term), we show that our maps mechanically exhibit  sparsity, in the sense that they have the ability to only impact adaptively $\bx$ on a subset of coordinates. We have proposed two simple generative models where this property helps estimation and applied this approach to high-dimensional single-cell datasets where we show, at a purely mechanical level, that we can recover meaningful maps. Many natural extensions of our work arise, starting with more informative sparsity-inducing norms (e.g., group lasso), and a more general approach leveraging the Bregman geometry for more ambitious $W_c$ problems, such as barycenters.

\bibliography{biblio}
\bibliographystyle{plainnat}

\clearpage
\appendix

\end{document}